\documentclass[fleqn,10pt]{wlscirep}% Nature Scientific Reports style

\usepackage[utf8]{inputenc}
\usepackage[T1]{fontenc}
\usepackage{float}

\usepackage{etoolbox}% Provides toggles
\newtoggle{arxiv}% Swap between arXiv and journal/conference version
\newcommand{\flexiref}[2]{\iftoggle{arxiv}{\autoref{#1}}{section \hyperref[#1]{\textit{#2}}}}% For referencing sections
\newcommand{\appref}[1]{\hyperref[#1]{Appendix~\ref*{#1}}}% For referencing subsections of the appendix
\newcommand{\mysection}[1]{\iftoggle{arxiv}{\section{#1}}{\section*{#1}}}
\newcommand{\mysubsection}[1]{\iftoggle{arxiv}{\subsection{#1}}{\subsection*{#1}}}
\newcommand{\mysubsubsection}[1]{\iftoggle{arxiv}{\subsubsection{#1}}{\subsubsection*{#1}}}
\newcommand{\citep}[1]{\cite{#1}}
\newcommand{\citet}[1]{\cite{#1}}

\usepackage{microtype}%
\usepackage{graphicx}%
\usepackage{multirow}%
\usepackage{amsmath,amssymb,amsfonts}%
\usepackage{amsthm}%
\usepackage{mathrsfs}%
\usepackage[title]{appendix}%
\usepackage{xcolor}%
\usepackage{textcomp}%
\usepackage{manyfoot}%
\usepackage{booktabs}%
\usepackage{algorithm}%
\usepackage{algorithmicx}%
\usepackage{algpseudocode}%
\usepackage{listings}%s
\usepackage{doi}%
\usepackage{subcaption}
\usepackage[capitalize]{cleveref}
\usepackage[detect-all]{siunitx}% Argument [detect-all] makes the font match the text

\DeclareSIUnit\ounce{oz}

\DeclareRobustCommand{\okina}{%
  \raisebox{\dimexpr\fontcharht\font`A-\height}{%
    \scalebox{0.8}{`}%
  }%
}
\DeclareUnicodeCharacter{02BB}{\okina}

\begin{document}

%\title[Article Title]{\textbf{A Multimodal Dataset of NEON Carabids}}
% \title{\textbf{A Multimodal Dataset of NEON Carabids [UPDATE?]}}

\title{A continental-scale dataset of ground beetles with high-resolution images and validated morphological trait measurements}

\author[1,\dag,*]{S~M~Rayeed}
\author[2,\dag,*]{Mridul~Khurana}
\author[3,\dag]{Alyson~East}

\author[4]{Isadora~E.~Fluck}
\author[5,8]{Elizabeth~G.~Campolongo}
\author[5]{Samuel~Stevens}

\author[6,7]{Iuliia~Zarubiieva}
\author[6]{Scott~C.~Lowe}

\author[9]{Michael~W.~Denslow}
\author[10]{Evan~D.~Donoso}
\author[5]{Jiaman~Wu}
\author[5]{Michelle~Ramirez}

\author[4]{Benjamin~Baiser}
\author[1]{Charles~V.~Stewart}

\author[11]{Paula~Mabee}
\author[5,8]{Tanya~Berger-Wolf}

\author[2]{Anuj~Karpatne}
\author[12]{Hilmar~Lapp}
\author[9]{Robert~P.~Guralnick}

\author[6,7,\ddag]{Graham~W.~Taylor}
\author[3,\ddag,*]{Sydne~Record}

\affil[1]{Rensselaer Polytechnic Institute, Department of Computer Science, Troy NY, 12180, USA}
\affil[2]{Virginia Tech, Department of Computer Science, Blacksburg VA, 24061, USA}
\affil[3]{The University of Maine, Department of Wildlife, Fisheries, and Conservation Biology, Orono ME, 04469, USA}
\affil[4]{University of Florida, Department of Wildlife Ecology and Conservation, Gainesville FL, 32611, USA}
\affil[5]{The Ohio State University, Department of Computer Science and Engineering, Columbus OH, 43210, USA}
\affil[6]{Vector Institute, Toronto ON, M5G 0C6, Canada}
\affil[7]{University of Guelph, School of Engineering, Guelph ON, N1G 2W1, Canada}
\affil[8]{The Ohio State University, Imageomics Institute \& ABC Global Climate Center, Columbus OH, 43210, USA} 

\affil[9]{University of Florida, Florida Museum of Natural History, Gainesville FL, 32611, USA}
\affil[10]{National Ecological Observatory Network, Pu'u Maka'ala Natural Area Reserve, Hilo HI, 96720, USA}
\affil[11]{Battelle, National Ecological Observatory Network, Boulder, CO 80301, USA}
\affil[12]{Duke University, Department of Biostatistics and Bioinformatics, Durham NC, 27708, USA}

\affil[$\dag$]{These authors contributed equally to this work: S M Rayeed, Mridul Khurana, and Alyson East}
\affil[$\ddag$]{These authors advised equally to this work: Graham W. Taylor and Sydne Record}
\affil[*]{Corresponding Authors: S M Rayeed (\href{mailto: rayees@rpi.edu}{rayees@rpi.edu}), Mridul Khurana (\href{mailto: mridul@vt.edu}{mridul@vt.edu}), Sydne Record (\href{mailto: sydne.record@maine.edu}
{sydne.record@maine.edu})}

% {\noindent \dag~Equal contribution.\quad
%  \ddag~Equal advising.\quad
%  * ~Corresponding author.}


% \author*[1,2]{\fnm{First} \sur{Author}}\email{iauthor@gmail.com}



% \affil*[1]{\orgdiv{Department}, \orgname{Organization}, \orgaddress{\street{Street}, \city{City}, \postcode{100190}, \state{State}, \country{Country}}}

% \affil[2]{\orgdiv{Department}, \orgname{Organization}, \orgaddress{\street{Street}, \city{City}, \postcode{10587}, \state{State}, \country{Country}}}

% \affil[3]{\orgdiv{Department}, \orgname{Organization}, \orgaddress{\street{Street}, \city{City}, \postcode{610101}, \state{State}, \country{Country}}}

\begin{abstract}
Despite the ecological significance of invertebrates, global trait databases remain heavily biased toward vertebrates and plants, limiting comprehensive ecological analyses of high-diversity groups like ground beetles. Ground beetles (\textit{Coleoptera: Carabidae}) serve as critical bioindicators of ecosystem health, providing valuable insights into biodiversity shifts driven by environmental changes. While the National Ecological Observatory Network (NEON) maintains an extensive collection of carabid specimens from across the United States, these primarily exist as physical collections, restricting widespread research access and large-scale analysis. To address these gaps, we present a multimodal dataset digitizing over 13,200 NEON carabids from 30 sites spanning the continental US and Hawaii through high-resolution imaging, enabling broader access and computational analysis. The dataset includes digitally measured elytra length and width of each specimen, establishing a foundation for automated trait extraction using AI. Validated against manual measurements, our digital trait extraction achieves sub-millimeter precision, ensuring reliability for ecological and computational studies. By addressing invertebrate under-representation in trait databases, this work supports AI-driven tools for automated species identification and trait-based research, fostering advancements in biodiversity monitoring and conservation.
\end{abstract}



\flushbottom

\maketitle

\thispagestyle{empty}

\mysection{Background and Summary}
\label{background}
Ground beetles (\textit{Coleoptera: Carabidae}, commonly known as \textit{carabids}), are among the most diverse and widespread invertebrate groups with over 40,000 known species \citep{lovei1996ecology}. Carabids serve as critical bioindicators of ecosystem health due to their sensitivity to environmental changes and functional roles within terrestrial ecosystems \citep{koivula2011useful, pearce2006use}. Their responses to habitat alteration, climate variability, and ecological disturbances provide critical insights into biodiversity trends and conservation priorities \citep{rainio2003ground, kotze2011forty, ghannem2018beetles, mcgeoch1998selection}, while their diverse feeding habits position them as key players in regulating pest populations, nutrient cycling, and maintaining soil ecosystem stability \citep{holland2002carabid, kromp1999carabid, holland2002small}. Their long evolutionary history \citep{grimaldi2005evolution} resulted in remarkable ecological and morphological diversity, with species adapting to virtually every terrestrial habitat from arctic tundra to tropical rainforests \citep{erwin2007taxonomic, erwin2004biodiversity}. Despite this morphological diversity, carabids lack systematic quantification of morphological traits that link directly to ecological processes. Furthermore, even with their ecological importance and ease of sampling using standardized pitfall trapping methods \citep{spence1997beetle}, significant data gaps limit trait-based research on carabids, creating barriers to understanding community assembly, ecosystem functioning, and species-level responses to global change.

Trait-based approaches are essential for understanding community assembly, ecosystem functioning, and responses to environmental change by linking measurable physiological, morphological, and behavioral characteristics to ecological processes \citep{mcgill2006rebuilding, violle2007let, cadotte2011beyond}. Traits such as body size, dispersal capacity, and habitat preference are highly sensitive to environmental stressors, enabling researchers to study niche partitioning, metabolic trade-offs, and species-level responses to global change \citep{chase2009ecological, brown2004toward}. However, global trait databases are heavily biased towards vertebrates and plants, with invertebrates largely underrepresented despite their dominant contribution to biodiversity \citep{etard2020global, guerra2020blind, cardoso2011seven}. This ``invertebrate gap'' limits comprehensive cross-taxon analyses and holistic ecological understanding, especially for hyper-diverse groups like carabids. Their complex taxonomy and subtle morphological differences, such as elytral striations or antennal segmentation, require labor-intensive and meticulous scrutiny for accurate species identification. The vast taxonomic diversity, the continuous discovery of new species \citep{kippenhan2012rediscovery}, and the scarcity of well-curated, high-quality image datasets hinder the development of automated identification tools \citep{blair2020robust, koivula2011useful, yang2022deep}. Even when available, carabid image datasets often lack the resolution needed for precise trait measurements, and no high-quality, fully documented datasets of carabid traits with corresponding publicly available images exist for North American species, limiting regional ecological studies \citep{blagoderov2012no, hedrick2020digitization}. These challenges have broader implications for ecological studies. For example, the degree of intraspecific variation across geographic clines, life stages, and environmental conditions remains unknown for most species \citep{ribera1999comparative, sukhodolskaya2016intra}. 

In an effort to overcome these challenges, we utilized data collected by the National Ecological Observatory Network (NEON), a continental-scale observatory spanning 81 sites across 20 ecoclimatic domains that provides standardized, long-term ecological data, including sampling of carabids  \citep{keller2008continental}. NEON’s trait datasets for plants, small mammals, and fish have driven significant advances in macroecological research, supporting studies on functional diversity, ecological theory testing, and predictive modeling of community responses to environmental change \citep{andersen2021engineered, read2018among, thibault2019seabirds, serbin2019arctic, astorga2025hantavirus, zatkos2021geophysical, fluck_influence_2024, mcgrew_abiotic_nodate}. These datasets have facilitated eco-evolutionary analyses and integration with remote sensing data \citep{jetz2019essential}. However, while NEON’s ground beetle monitoring program collects extensive carabid specimens using standardized pitfall trapping, it does not include comprehensive specimen images or trait measurements \citep{hoekman2017design, groundbeetlesequencesDNAbarcode}. Previous NEON carabid studies have made important contributions through species richness and DNA barcoding analyses \citep{hoekman2017design, groundbeetlesequencesDNAbarcode}, and recent efforts have demonstrated the potential of image-based approaches through computer vision for species identification \citep{blair2020robust} and body size measurements of invertebrate bycatch from NEON pitfall traps \citep{kaspari_geographic_2024}. These studies highlight the value of computational approaches to morphological analysis in ecological research. To fully realize the potential of such approaches and support reproducible research, there is a growing need for FAIR-compliant (Findable, Accessible, Interoperable, Reusable) \citep{wilkinson2016fair} datasets that provide both high-resolution images and standardized trait measurements with transparent methodologies. Our work addresses this gap by providing publicly available images of NEON carabids with accompanying validated trait measurements, enabling broader access for computational analysis and methodological development.

In the work we present here, we curated a multimodal dataset of over 13,200 NEON carabid specimens, integrating high-resolution imaging with digital measurements of elytral length and width --- key morphological traits critical for species identification and functional analysis. These traits correlate with dispersal capacity, habitat preference, and responses to environmental stressors, making them important candidates for better understanding the ecological role of carabids \citep{Shegelski2019, Stoces2025, desroches2023trophic, homburg2013broad, gaston2013macroecological, kotze2000colonization, ribera_effect_2001, liu2025beetleflow}. Unlike existing datasets \citep{Steinke2024,bioscan5m,orsholm2025,mehrab2025fish}, our high-resolution imaging follows rigorous protocols, ensuring reproducible results suitable for both human and computational analysis \citep{east2025optimizing, keller2023ten}. We validated digital measurements against manual caliper-based measurements, achieving sub-millimeter precision and establishing transparent error rates, addressing a common limitation in trait databases where validation is often unreported \citep{perez2019botanic, gallagher2020open, gallagher2023open}. This dataset is the first to provide high-quality images and trait measurements for North American carabids, overcoming the limitations of low-resolution, poorly documented, or region-specific datasets.

Furthermore, we integrate morphological data with NEON’s extensive environmental and climatic datasets, enabling researchers to investigate trait-environment interactions across diverse ecosystems. These data support studies of ecological phenomena such as niche differentiation, energy allocation strategies, and adaptive responses to environmental change at unprecedented spatial scales \citep{chase2009ecological, brown2004toward, mcgill2006rebuilding}. By quantifying intraspecific variation in carabid traits, our dataset complements NEON’s plant, mammal, and fish datasets, facilitating cross-taxon analyses of functional diversity, trophic interactions, and ecosystem stability \citep{etard2020global, schneider2022getting}. Its standardized format and comprehensive metadata ensure integration with global biodiversity platforms, enhancing accessibility and reusability \citep{edwards2000interoperability, wilkinson2016fair}. 

This dataset provides immediate value for both ecological research and computational applications. The high-resolution images and validated measurements serve as a robust foundation for interdisciplinary research at the intersection of computer vision and biology, enabling applications such as automated trait extraction, species identification \citep{gaston2004automated, stevens2024bioclip,rayeed2025beetleverse,gong2025clibd,rayeed2025fine,maruf2024vlm4bio}, and discovery of novel morphological patterns \citep{elhamod2023discovering, Liu_2025_ICCV, khurana2024hierarchical}. For ecological applications, the dataset addresses a critical need for standardized trait measurements with a validated digital approach achieving sub-millimeter precision with transparent error quantification. Such precision is particularly valuable in ecological studies, where datasets are typically small and heterogeneous, limiting the effectiveness of conventional computer vision approaches that require large, standardized training sets \citep{todman2023small, stevens2025mind}. By adhering to FAIR data principles and integrating with NEON’s infrastructure, we ensure its long-term preservation and discoverability. \citep{wilkinson2016fair, nelson2019history}. This dataset has the potential to contribute to applications that include real-time environmental monitoring, conservation planning, ecological forecasting, and the development of scalable tools for biodiversity assessment, bridging traditional morphological approaches with emerging computational methods to address the challenges of rapid environmental change \citep{urban2016improving, bush2017connecting}.

\mysection{Methods}
\label{methods}
In this section, we describe the standardized protocols and procedures used to collect, process, image, measure, and analyze the beetle data across NEON terrestrial sites. We detail the experimental design for pitfall trap sampling, imaging of specimens, individual specimen segmentation, and automated trait measurement. All methods are fully described to ensure reproducibility, with specific data inputs referenced using the appropriate data citation format, following the guidelines for reproducible methods.

\mysubsection{Data Collection}

\mysubsubsection{Sample Sites and Schema}
\label{neon}
NEON operates 81 field sites in the United States of America, including 47 terrestrial and 34 aquatic sites distributed across 20 ecoclimatic domains. These sites are strategically distributed to capture ecological variation across diverse ecosystems, from the continental United States to Hawaii and Puerto Rico. Domains are statistically defined by climate, vegetation, landforms, and ecosystem dynamics into 20 representative regions \citep{keller2008continental}. As a result, NEON sites encompass diverse ecosystems including deciduous forests, grasslands, shrublands, taiga, and desert biomes, providing a comprehensive cross-section of North American ecological conditions. The observatory completed construction in 2019, which is the first year of full operations across all 81 sites. NEON’s terrestrial data collection integrates organismal abundance, diversity and trait data on seminal taxa (Ground Beetles, Small Mammals, Birds, Plants, Ticks, and Mosquitoes), as well as a variety of automated environmental sensors and fine-scale remote sensing data harmonized. The collocation of sampling effort across taxa and environmental gradients presents unprecedented opportunity for inference both on beetles in the environment and cross taxa biodiversity assessments. 

\begin{figure}[!t]
\centering
\includegraphics[width=\textwidth]{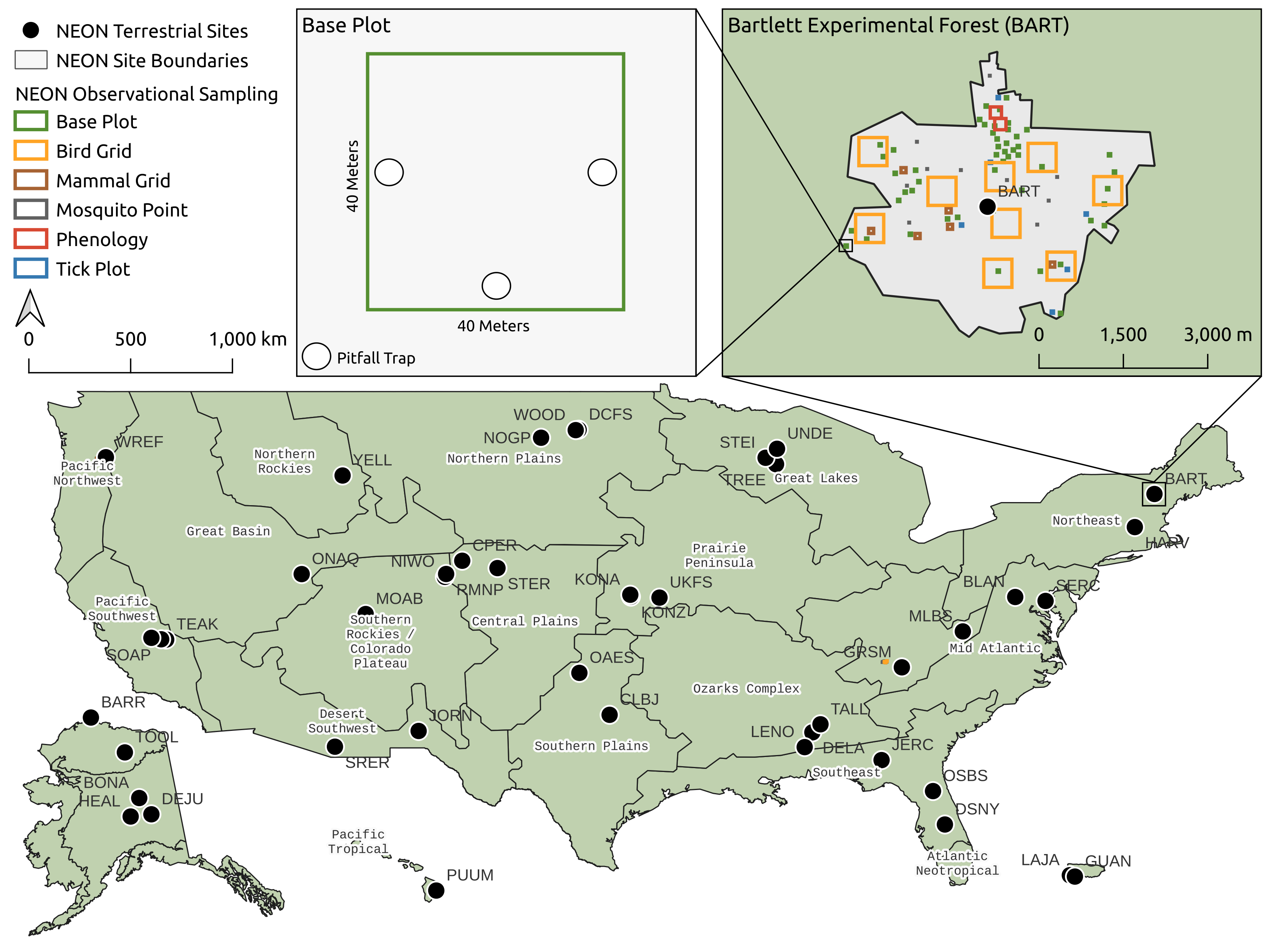}
\captionsetup{justification=justified}
\caption{Map of NEON terrestrial sites across the United States of America. Each circle marks a NEON terrestrial site: black fill indicates sites where we imaged carabid samples. The inset shows the Bartlett Experimental Forest BART site in detail, illustrating NEON’s spatial sampling design across plots designated for specific observation types (e.g., birds, mammals, phenology), carabids are collected at (green) Base Plot. The ``Base Plot'' diagram (upper middle) outlines the standardized pitfall trap configuration used for ground beetle sampling, with three traps positioned at the south, east, and west edges respectively of a \qtyproduct{40 x 40}{\meter} plot.}
\label{fig:StudyArea}
\end{figure}

\mysubsubsection{Ground Beetle Collections from Pitfall Traps at NEON}\label{pitfall}
Each terrestrial NEON site implements standardized ground beetle collection protocols using pitfall traps. The sampling follows a spatially balanced stratified random design, with sites initially containing 10 distributed plots (reduced to 6 plots starting in 2023). Within each plot, pitfall traps are positioned at the cardinal points (south, east, and west) of a \qtyproduct{40 x 40}{\meter} area within designated sampling sites. The northern position was discontinued in 2018, reducing the configuration from four to three traps per plot \citep{li_standardized_2022}. Trapping occurs during the growing season (typically May through October, with regional climate variations), with consistent sampling effort maintained across all sites. Ground pitfall traps are collected at two-week intervals, with up to 13 collection bouts per year. The traps consist of \SI{16}{\ounce} deli containers filled with \SIrange{150}{250}{\milli\litre} of propylene glycol, which serves as both a killing agent and preservative. Each trap deployment lasts approximately 14 days, with precise set and collection dates recorded. Field technicians document trap condition (i.e., cup status, lid status, fluid level) during collection and transfer specimens to 95\% ethanol within 24 hours.

\begin{figure}[!t]
\centering
\includegraphics[width=\textwidth]{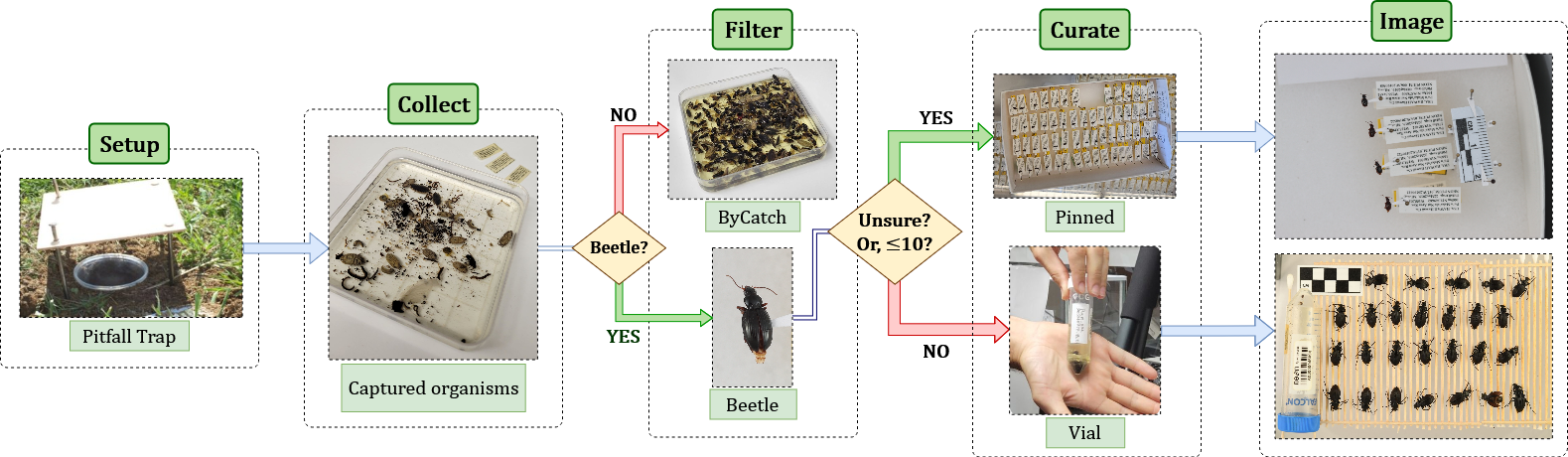}
\captionsetup{justification=justified}
\caption{Workflow for standardized ground beetle collection and processing at NEON sites. The process begins with pitfall trap setup \citep{GroundBeetles} and proceeds through the collection of captured organisms. Specimens are filtered to separate carabids from bycatch. If identified as a beetle, the specimen is further curated based on identification confidence and quantity: specimens with uncertain ID or fewer than ten per species are pinned; otherwise, they are preserved in ethanol (vial).}
\label{fig:pitfall-beetle-collection}
\end{figure}

Laboratory processing involves sorting trap contents to separate carabids from other captured organisms (vertebrate and invertebrate bycatch). NEON parataxonomists then identify carabids to species or morphospecies level. The preservation method for specimens varies based on identification confidence and specimen quantity. If species identification is uncertain, the specimen is pinned and sent for expert taxonomist review (hereafter referred to as ``pinned specimens''). For confirmed identifications, if ten or fewer individuals of a species are collected in a sampling bout, all are pinned; if more than ten are collected, all are preserved in ethanol (hereafter referred to as ``vial specimens''). Figure~\ref{fig:pitfall-beetle-collection} shows a graphical overview of the data collection and imaging workflow.

The comprehensive data management system includes detailed records of collection conditions, taxonomic identifications, and archiving information \cite{DP1.10022.001_RELEASE2025}. Quality control measures are implemented throughout the process, with taxonomic identifications following established references \citep{lorenz2005systematic, bousquet2012catalogue}. This standardized methodology enables researchers to investigate beetle populations and communities across diverse ecosystems for ecological analysis, indicator species analysis, habitat assessment, and conservation purposes, with additional value through integration with other NEON measurements at co-located sampling points.

In this study, we worked with pinned and vial specimens. Vial specimens were obtained upon request from the NEON Biorepository for the sample year 2018 \citep{Portal2022-ho, Portal2022-qu}. Pinned specimens were photographed at the NEON office located at the Puʻu Makaʻala (PUUM) Natural Area Reserve in Hilo, Hawaii \citep{NEON-pinned-specimens, NEON-pinned-beetles-metadata}. At the PUUM reserve, there are two distinct collections of pinned beetles: one at the on-site facility where all specimens across sample years were imaged and presented in this work; and another at the Bishop Museum in Honolulu, Hawaii.

\begin{figure}[!t]
\centering
\includegraphics[width=\textwidth]{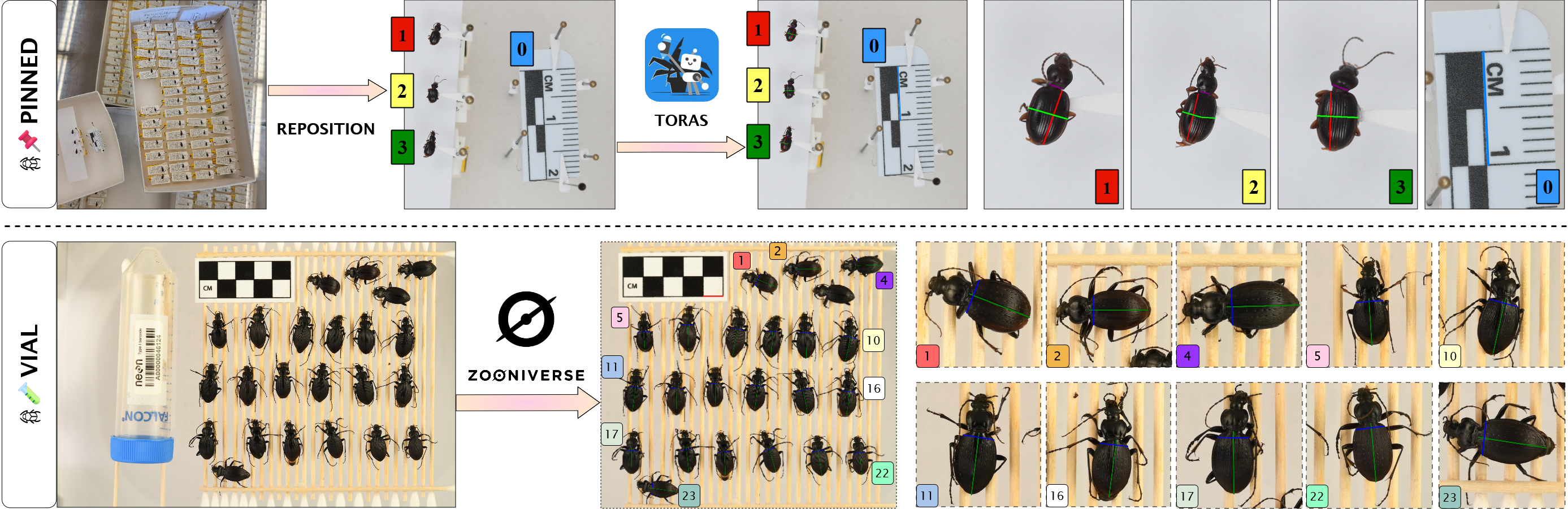}
\captionsetup{justification=justified}
\caption{Workflow for standardized data collection protocol for beetles across two specimen types: pinned (top) and vial (bottom). \textbf{Top (Pinned Workflow)}: Pinned specimens, first repositioned and imaged, are being processed using TORAS for precise trait measurements of elytral length (red line), basal pronotum width (purple), maximum elytral width (green), and scale bar (blue). \textbf{Bottom (Vial Workflow)}: Ethanol-preserved beetles, arranged on gridded boards and imaged, are uploaded to the Notes from Nature project, where individual beetles are annotated and labeled for measuring elytral length (green) and width (blue). Enlarged depictions of trait annotations on fictional individual beetles are provided in \cref{fig:Fictional-Traits} to illustrate the measurement locations more clearly.}
\label{fig:traits-workflow-3}
\end{figure}

\mysubsubsection{Imaging NEON Specimens}\label{imaging-beetles}
Beetle specimens were photographed using a standardized imaging station optimized for consistent, high-resolution capture. Vial and pinned specimens were imaged separately, each with equipment and settings tailored to specimen size and presentation (Table \ref{tab:imaging-setup}). These high-quality, standardized images ensure accurate taxonomic assessment and support the development of a reliable digital reference collection for NEON beetle specimens \citep{east2025optimizing}. The overall workflow for imaging and subsequent trait measurement across both specimen types is illustrated in Figure \ref{fig:traits-workflow-3}.

\paragraph{Equipment Setup and General Protocol.} Cameras were mounted on a copy stand or tripod to ensure stability and repeatability across imaging sessions. The imaging stations were calibrated before each session to maintain consistency in focal distance, lighting, and specimen positioning. Lighting was provided by two flashes positioned on either side of the subject to produce even illumination across the entire field of view. To minimize glare from reflective insect cuticles and laminated labels, light was diffused using a shadow box, which produced soft, uniform lighting and preserved fine morphological detail critical for taxonomic identification. All images were acquired using consistent manual settings, with exposure, ISO, and white balance optimized for each imaging context but standardized within datasets.

\paragraph{Imaging Protocol: Pinned Specimens.} Pinned specimens required an approach to accommodate their mounted presentation and associated identification labels. These specimens were imaged in groups of 3--5 beetles per frame, deliberately arranged on the right side of the image area in standard collection boxes with a white background. This arrangement optimized frame utilization while ensuring sufficient resolution for each individual specimen. Pins were inserted at uniform heights for all specimens (including the scale bar pin) to ensure a singular focal plane across the entire frame, effectively eliminating the need for focus stacking and simplifying the imaging workflow. This standardization of pin heights proved crucial for maintaining consistent focus across specimens of varying sizes and morphologies. Each pinned specimen frame included a scale bar placed in the upper left corner, mounted at precisely the same height as the specimens to ensure accurate size referencing. All specimens were photographed in standardized dorsal orientation to maximize the visibility of taxonomically relevant features, though in select cases where ventral or lateral diagnostic characteristics were particularly significant, additional images were captured from these perspectives. Each pinned specimen was presented with its complete suite of identification labels, including metadata tags, DNA barcoding information, parataxonomist determinations, expert taxonomist verifications, and any correction or update labels documenting taxonomic revisions over time. These labels were rotated such that the text appears upside down in the images (see \cref{fig:pitfall-beetle-collection}). This orientation was chosen because the specimens are very small, and textual elements in the background could otherwise interfere with downstream AI tasks, such as segmentation of individual beetles or their body parts.

This strategic label positioning minimized text-based visual distractions in the immediate vicinity of the specimens, facilitating more accurate computer vision analysis while still preserving the critical metadata within the same image. This comprehensive inclusion of label data directly in the images created a complete digital record of both morphological and metadata elements associated with each specimen. When necessary for additional clarity, labels were carefully repositioned to avoid obscuring important specimen traits, though care was taken to document the original label positioning before manipulation. The Canon EOS 7D with \SIrange{24}{105}{\milli\metre} macro lens was employed exclusively for pinned specimens, using diffused flash lighting positioned on opposite sides of the specimens. The tracing paper diffusion cylinders were particularly important for these specimens to reduce reflections from the highly reflective cuticle surfaces while maintaining detail visibility. This lighting approach proved essential for capturing fine structural details on the often-glossy exoskeletons while minimizing harsh shadows that might obscure important taxonomic features. Careful attention was paid to proper white balance calibration to ensure accurate color reproduction, particularly important for distinguishing subtle coloration that might represent taxonomically significant traits or variation within species groups.

\paragraph{Imaging Protocol: Vial Specimens.} Vial specimens required specialized handling due to their preservation method. These specimens were carefully removed from ethanol preservation and allowed to dry completely on absorbent paper before imaging to prevent glare and reflections from wet surfaces. After sufficient drying time, specimens were arranged systematically on wooden stakes positioned in parallel across the frame, providing elevation and structure to ensure beetles remained in a consistent dorsal view position for optimal elytra visibility and posterior measurement. The wooden stakes served as supports, preventing specimens from resting directly on the imaging surface and establishing a consistent focal plane across all specimens within the frame. All specimens from a single vial were placed in one image to maintain collection integrity and simplify metadata association, allowing for efficient batch processing during subsequent analysis stages. Each vial specimen image included the original storage vial with its barcode prominently positioned, preferably in the left section of the frame for direct specimen-to-metadata linkage. This visual connection between specimens and their source container facilitated efficient data management and reduced the potential for misidentification or metadata dissociation during digital processing. A standardized scale bar was consistently placed in the upper left corner of the frame for accurate pixel-to-millimeter conversion, enabling precise morphometric measurements from the digital images. Specimens were meticulously positioned to maximize visibility of diagnostic morphological features, particularly pronotum shape, elytral texture, and other taxonomically significant structures. To prevent glare from reflective structures and ensure uniform lighting, a shadowbox was placed around the sample prior to imaging. The NIKON D500 camera was operated at 1/13s shutter speed with an f8.0 aperture and ISO\,100 to achieve maximum depth of field while maintaining image clarity. When required, the distance between the camera and the specimens was adjusted according to the size of the individuals and the number of beetles per vial, in order to optimize image framing and ensure consistent capture quality across the digital collection.

\paragraph{Image Quality Control and Processing.} Each imaging session began with calibration shots and concluded with thorough quality control procedures to ensure consistency and high image quality across the entire dataset. Quality control assessments included checking that each specimen was in focus, accomplished by verifying that key diagnostic features were clearly visible on every individual within the frame. Lighting quality was rigorously evaluated to verify adequate illumination with no harsh shadows or blown-out highlights that might obscure important morphological details. Technicians verified the visibility of all relevant taxonomic dorsal features specified by consulting entomologists, ensuring that images would serve their intended identification and research purposes. Label legibility was confirmed for pinned specimens, with particular attention to barcodes and handwritten annotations that might be challenging to digitize through other means. Proper inclusion and positioning of scale bars was verified in each image to enable accurate measurements, and correct white balance and color fidelity were assessed against physical color standards included in calibration shots. Images failing any quality control criterion were retaken immediately, while the specimens remained in their arranged positions. All images passing quality control were backed up in triplicate before any post-processing occurred, preserving the raw documentation for archival purposes. Initial processing included only non-destructive adjustments to exposure and contrast when necessary to improve feature visibility, with all processing steps documented in accompanying metadata files. This rigorous imaging and quality control protocol resulted in a comprehensive digital collection that accurately represented the physical specimens, facilitating both current research applications and future reference needs. The standardization of imaging approaches between the vial and pinned specimen datasets, despite their different handling requirements, created a cohesive digital collection that maintains consistency while accommodating the unique characteristics of each preservation method.

\begin{table}[!t]
\centering
\resizebox{\textwidth}{!}{%
\begin{tabular}{lll}
\toprule
\textbf{Parameter} & \textbf{Vial Dataset} & \textbf{Pinned Dataset} \\
\midrule
Camera Body & NIKON D500 & Canon EOS DSLR (model 7D) \\
Lens & \SI{60}{\milli\metre} f/2.8G ED macro lens & \SIrange{24}{105}{\milli\metre} macro lens \\
Lighting & 
Shadowbox (with lamps in the lower left and right)
& 2 flashes \\
Flash Positioning & No flash & Opposite, diffused \\
Diffusion Method & Rosco Tough white tracing paper between flash and fragment & White tracing paper cylinder between flash and tray\\
Shutter Speed & 1/13s & 1/100s \\
Aperture (f-stop) & f8.0 & f8.0 \\
ISO & 100 & 100 \\
White Balance & Auto & Auto \\
Background & Wooden dorsal alignment support & White standard collection tray \\
Scale & In the upper left corner on wooden props& In the upper left corner at pin height\\
Labels & Vial with barcode in the left& Under the pinned beetles (rotated for segmentation)\\
Mounting & Individuals dried and arranged on wooden support & Pinned specimens mounted dorsally \\
Orientation & Dorsal & Dorsal \\
\bottomrule
\end{tabular}
}
\captionsetup{justification=justified}
\caption{Summary of imaging setup for Vial and Pinned specimen datasets.}
\label{tab:imaging-setup}
\end{table}

\mysubsection{Trait Measurements}\label{trait-measurement}
Morphological traits, especially body size, represent fundamental dimensions of ecological variation that influence virtually all aspects of an organism's life history, physiology, and ecological interactions \citep{mcgill2006rebuilding, suding2008scaling}. For carabid beetles, elytral length and width serve as reliable proxies for body size \citep{chiari_monitoring_2014}, a core trait that links directly to metabolic theory\citep{enquist_scaling_2003}, microhabitat or disturbance adaptations\citep{ribera_effect_2001}\citep{Stoces2025}, dispersal capacity\citep{juliano_food_1986}, and competitive interactions\citep{brown2004toward}, making it valuable for understanding ecological processes and responses to environmental change.

\begin{figure}[!t]
\centering
\includegraphics[width=\textwidth]{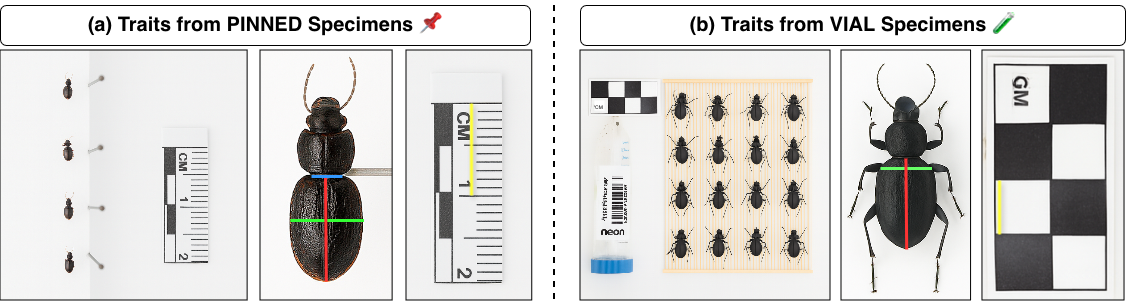}
\captionsetup{justification=justified}
\caption{Artificial depiction of trait measurement ``recipes'' for pinned and vial specimens. \textbf{(a) Pinned specimens}: left: Fictional group image of pinned beetles with scalebar, middle: one individual specimen with three traits measured (elytral length in red, basal pronotum width in blue, and maximum elytral width in green), right: measurement of the centimeter scalebar; \textbf{(b) Vial specimens}: left: Fictional group image of vial beetles with checkbox, middle: one individual specimen with two traits measured (elytral length in red, and elytral width in green), right: measurement of the centimeter checkerbox. \textit{Images shown in this figure were generated using Microsoft CoPilot.}}
\label{fig:Fictional-Traits}
\end{figure}

\mysubsubsection{Trait Measurements via Toronto Annotation Suite (TORAS)}\label{toras}
The pinned specimens were digitally measured using the Toronto Annotation Suite (TORAS)\cite{torontoannotsuite}, an AI-powered annotation platform that enables precise polygon and polyline drawing with human-in-the-loop machine learning assistance. For each specimen, we annotated three key morphological traits by drawing polylines: (1) \textbf{elytral length}, measured from the midpoint of the elytro-pronotal suture (the junction between the pronotum and elytra) to the midpoint of the elytral apex (posterior tip of the elytra); (2) \textbf{basal pronotum width}, measured as the width of the pronotum at its base, at the junction between the pronotum and elytra; and (3) \textbf{maximum elytral width}, measured as the greatest transverse distance (distance between the widest points) across both elytra, ideally orthogonal to the elytral length axis (see \cref{fig:Fictional-Traits}a).

The TORAS interactive segmentation tools, which combine state-of-the-art computer vision models with precise polygon/polyline editing capabilities, enabled accurate and consistent measurements across specimens. The platform's collaborative repository system allowed multiple researchers to contribute to the measurement process while maintaining annotation consistency through predefined ``recipes'' --- standardized annotation protocols that specified exact measurement landmarks for each trait. This approach minimized inter-observer variability while significantly reducing the time required for measurements compared to traditional manual methods. Each measurement specified not only start and end landmarks but also resolution guidelines (e.g., number of points per polyline), handling of suboptimal specimens (e.g., with open elytra), and a required order for annotations (e.g., the last measurement should always be the scale bar). 
For example, in the case of an open elytra, the polyline for maximum elytral width consisted of three segments: the left elytron, the inter-elytral gap along the dorsal midline, and the right elytron. For the measurement, only the first and third segments were included in the final measurement. Annotators were provided with visual examples and a video tutorial to ensure consistency. Once annotation masks were complete, we used a simple Python script to convert polyline lengths from pixels to millimeters, using the scale bar as a reference.

\mysubsubsection{Trait Measurements via Notes from Nature} \label{zooniverse}
To support large-scale morphological data collection from vial specimens, we collected trait annotations from 577 group images (using the Notes from Nature platform \citep{hill2012notes}), each containing multiple beetles placed on a wooden frame with a scale bar. The original group images were photographed at a high resolution of 5568~×~3712~pixels, exceeding the 1~MB per-image upload limit imposed by Notes from Nature. Therefore, we generated resized versions of the group images for annotation, and all trait measurements were performed on these resized images. For each image, we measured two traits: (1) \textbf{elytral length}, following the same definition as described in TORAS pipeline; and (2) \textbf{elytral width}, measured orthogonal to the elytral length axis as the transverse distance between the lateral margins of the elytra at the level of the mesothoracic legs (the first pair of legs extending beneath the elytra) (see \cref{fig:Fictional-Traits}b). To ensure consistency across annotators, we provided a standardized annotation protocol, closely aligned with that of the TORAS pipeline, that outlined anatomical landmarks, measurement conventions, and guidance for challenging cases, such as measuring damaged specimens. 

To evaluate measurement reliability, a subset of specimens was annotated independently by up to three contributors, allowing us to evaluate inter-annotator agreement (see \flexiref{sec:inter-annotator}{Vial Specimens: Inter-annotator agreement}). For elytral length, inter-annotator agreement was high, with low variability across measurements. Since this is a straightforward linear trait measured along the beetle’s midline, contributors were generally able to identify the correct landmarks even without entomological expertise. In contrast, elytral width showed greater variability, as the mesothoracic leg landmarks were sometimes partially occluded, making consistent measurement difficult. Consequently, vial-based elytral width measurements were excluded from the final analyses; however, both length and width data remain available in the dataset to support future ecological and morphological studies. Elytral width measurements, however, showed greater variability. The mesothoracic legs were sometimes partially occluded, making it difficult to identify their positions consistently. Due to this inconsistency, we excluded elytral width measurements of vial-specimens from our final analyses. However, both elytral length and width values are included in the dataset to support future studies that may benefit from broader ecological research applications.

\begin{figure}[!t]
\centering
\includegraphics[width=\textwidth]{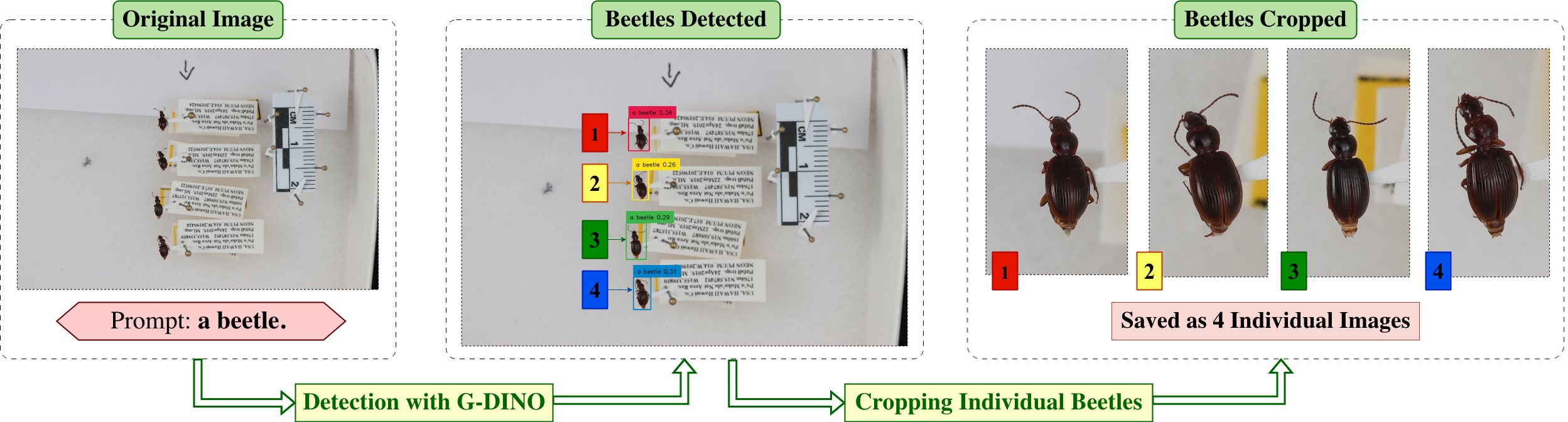}
\captionsetup{justification=justified}
\caption{Overview of the individual specimen segmentation workflow. \textbf{Left:} the original group image is processed with Grounding DINO using the prompt ``{\small\texttt{a beetle.}}'' \textbf{Middle}: detected beetles are labeled with bounding boxes. \textbf{Right:} each beetle is cropped and saved as an individual image after manual review and (potential) correction of the bounding boxes.}
\label{fig:beetle-detection}
\end{figure}

\mysubsection{Individual Specimen Segmentation} \label{individual_beetles}
To isolate individual specimens from high-resolution group images, we implemented a hybrid workflow that combines cutting-edge computer vision with human inspection (See \cref{fig:beetle-detection} for the segmentation workflow). First, images were processed with the open-vocabulary detector Grounding DINO \citep{liu2024grounding} --- an open-vocabulary object detector with strong zero-shot capabilities, well-suited for handling the diverse shapes and color patterns of beetle specimens. It produced an initial set of bounding boxes, rapidly converting thousands of group images into cropped candidate specimens.

Given the limitations of automated detection --- such as missing partially obscured individuals or drawing overly tight or loose bounding boxes --- we imported all candidate annotations into CVAT (\underline{C}omputer \underline{V}ision \underline{A}nnotation \underline{T}ool) \citep{cvat2025}, an interactive data annotation platform. In CVAT, every individual specimen image was reviewed: missing specimens were manually added, bounding boxes were adjusted to fully include anatomical features like antennae, and false positives were removed. The resulting manually annotated bounding box extractions were used to generate the final set of individual specimen images.

\mysection{Data Records}\label{records}

\noindent All images and trait measurement data, along with associated NEON metadata for pinned and vial specimens are publicly available on HuggingFace datasets hub, respectively at \href{https://huggingface.co/datasets/imageomics/Hawaii-beetles}{\small\texttt{imageomics/Hawaii-beetles}} and \href{https://huggingface.co/datasets/imageomics/2018-NEON-beetles}{\small\texttt{imageomics/2018-NEON-beetles}}. 
The {\small\texttt{Hawaii-beetles}} repository contains high-resolution group images of \textit{pinned specimens}, their corresponding individually cropped specimen images, along with trait measurements using TORAS \cite{NEON-pinned-beetles-metadata, NEON-pinned-specimens, rayeed2025HawaiiBeetles}. The images and trait measurements are released under a \href{https://creativecommons.org/licenses/by/4.0/}{{CC BY 4.0}} license, as shown in Table~\ref{tab:hawaii-beetles}. The {\small\texttt{2018-NEON-beetles}} repository contains group images of \textit{vial specimens} photographed at scale, along with resized versions of the group images, individual specimen crops derived from the resized group images, and corresponding trait measurements obtained from these resized images \cite{Fluck2018_NEON_Beetle, Portal2022-ho, Portal2022-qu}. All images and trait data are released under a \href{https://creativecommons.org/licenses/by-sa/4.0/}{{CC BY-SA 4.0}} license to match the original usage terms, as mentioned in Table~\ref{tab:2018neonbeetles}.

\mysubsection{Data Records of Pinned Specimens}\label{sec:pinned-records}

\paragraph{Directory layout.}
The repository follows a flat directory structure summarized in Table~\ref{tab:hawaii-beetles}. It includes the original group images of \textit{pinned specimens} ({\small\texttt{group\_images/}}), individually cropped specimen images ({\small\texttt{individual\_specimens/}}), and corresponding metadata ({\small\texttt{images\_metadata.csv}}) and trait measurement ({\small\texttt{trait\_annotations.csv}}) files in CSV format. The metadata file links each cropped specimen image to its parent group image and the associated NEON identifier ({\small\texttt{individualID}}). All images are in PNG format, and documentation with usage notes are included in the {\small\texttt{README.md}} file.

\begin{table}[ht]
\centering
\begin{tabular}{llrl}
\toprule
\textbf{Folder / file} & \textbf{Content} & \textbf{\# records} & \textbf{Licence}\\
\midrule
{\small\texttt{group\_images/}} & Original group images & \num{162} & CC-BY-4.0\\
{\small\texttt{individual\_specimens/}} & Cropped images of individual pinned specimens & \num{1614} & CC-BY-4.0\\
{\small\texttt{images\_metadata.csv}} & Specimen-level metadata & \num{1614} & CC-BY-4.0\\
{\small\texttt{trait\_annotations.csv}} & Specimen-level trait measurements & \num{1579} & CC-BY-4.0\\
{\small\texttt{README.md}} & Dataset card and usage notes & — & —\\
\bottomrule
\end{tabular}
\caption{Directory structure of the \emph{pinned specimens} repository: \href{https://huggingface.co/datasets/imageomics/Hawaii-beetles}{\small\texttt{imageomics/Hawaii-beetles}}.}
\label{tab:hawaii-beetles}
\end{table}

\paragraph{Image naming.} 
A systematic naming convention was adopted to ensure a clear link between specimen images and their associated metadata. Original group images, referenced by {\small\texttt{groupImageFilePath}} in the metadata, retain their camera-assigned filenames in the format {\small\texttt{group\_images/IMG\_<id>.png}} (where {\small\texttt{<id>}} corresponds to the image’s original camera roll identifier), preserving the original identifiers assigned during image capture. Individually cropped specimens derived from a given group image are referenced by {\small\texttt{individualImageFilePath}} and are structured as \\
\centerline{\small\texttt{individual\_specimens/IMG\_<id>\_specimen\_<number>\_<taxonID>\_<individualID>.png}}

In this format, the {\small\texttt{IMG\_<id>}} component directly corresponds to the parent group image, while {\small\texttt{<number>}} indicates integer index of the beetle's specific position within that group image. The {\small\texttt{<taxonID>}} is a six-letter code derived from the first three letters of the genus followed by the first three letters of the species epithet.
The {\small\texttt{<individualID>}} is a unique specimen identifier for linking to external NEON data products, it begins with the prefix {\small\texttt{NEON\.BET.D20}} (indicating NEON Domain 20), followed by six digits. This structured convention creates a robust linkage between the physical specimen, its high-resolution images, and the metadata records.

\paragraph{Metadata.} 
Specimen-level metadata is provided in the CSV file {\small\texttt{images\_metadata.csv}} to link image files to their ecological and taxonomic context. Each record/row in the file includes {\small\texttt{groupImageFilePath}}, {\small\texttt{individualImageFilePath}}, {\small\texttt{individualID}}, {\small\texttt{taxonID}}, {\small\texttt{ScientificName}}, {\small\texttt{plotID}}, {\small\texttt{trapID}}, {\small\texttt{plotTrapID}}, {\small\texttt{collectDate}}, {\small\texttt{ownerInstitutionCode}}, {\small\texttt{catalogNumber}}. The {\small\texttt{individualImageFilePath}} provides the path to all individually cropped specimen images, and the {\small\texttt{groupImageFilePath}} maps to their corresponding group/tray images. Other fields provide essential information on the specimen's identity, collection location within the NEON site, and the date of collection. 

Taxonomic information is provided through the {\small\texttt{taxonID}}, a six-letter code, and {\small\texttt{ScientificName}} provides the full binomial scientific name containing \textit{Genus} and \textit{species} epithet. The {\small\texttt{plotID}} specifies the NEON plot where the beetle was collected, while the {\small\texttt{trapID}} indicates the cardinal direction (E, S, or W) of the pitfall trap within that plot; both combined form the {\small\texttt{plotTrapID}}. The {\small\texttt{collectDate}} field records the date of collection in a {\small\texttt{YYYYMMDD}} format. To ensure traceability to the physical specimens in NEON's biorepository, the metadata includes the {\small\texttt{ownerInstitutionCode}} and the {\small\texttt{catalogNumber}}, which are NEON identifiers. 

Furthermore, each record is associated with a unique specimen identifier {\small\texttt{individualID}}, which serves as the primary key linking the image and trait data to the broader NEON database. Through this identifier, researchers can retrieve additional contextual information from NEON’s data portal, including environmental variables, plot and trap IDs, collection metadata, and, for some specimens, associated DNA barcode data.

\paragraph{Annotations and Trait Measurements.}  
All morphological trait measurements are consolidated into a single CSV file, {\small\texttt{trait\_annotations.csv}}.  Each row corresponds to a single measured specimen, providing complete measurements of beetle traits. Traits are defined by landmark point pairs, with each record containing all data from the initial pixel coordinates to the physical measurements. Each row links to its source group image via {\small\texttt{groupImageFilePath}} and records the {\small\texttt{BeetlePosition}} of the specimen within that image. The landmark coordinates are provided as $(x,y)$ pixel coordinate pairs following a top-left image origin ($x$ increasing rightward, $y$ downward). For each landmark pair, the corresponding pixel-distance field, {\small\texttt{px\_<trait>}}, gives the Euclidean distance between its endpoints. We calculate the per-image calibration factor, $k$ (in pixels per centimeter) as {\small$k = \texttt{px\_scalebar} / \texttt{cm\_scalebar}$}, where {\small\texttt{cm\_scalebar}} is fixed to \SI{1.0}{\centi\metre}. Physical trait measurements are provided in the {\small\texttt{cm\_<trait>}} fields, which are derived using the formula:
{\small$\texttt{cm\_<trait>} \;=\; \texttt{px\_<trait>} / k$}.
This yields the final scale-corrected traits: maximum elytral length ({\small\texttt{cm\_elytra\_max\_length}}), basal pronotal width ({\small\texttt{cm\_basal\_pronotum\_width}}), and maximum elytral width ({\small\texttt{cm\_elytra\_max\_width}}).

\mysubsection{Data Records of Vial Specimens}\label{sec:vial-records}

\paragraph{Directory layout.}
The repository structure for the \emph{vial specimens} is summarized in Table~\ref{tab:2018neonbeetles}. It contains the original group images ({\small\texttt{group\_images/}}), their resized counterparts used for annotation ({\small\texttt{group\_images\_resized/}}), individually segmented beetle images ({\small\texttt{individual\_specimens/}}), and corresponding metadata and trait measurement files in CSV format. In addition, there is a dedicated subdirectory that provides pre-organized training and testing subsets derived from the individual segmentation {\small\texttt{Separate\_segmented\_train\_test\_splits\_80\_20/}}. All images and derived data are distributed under the \href{https://creativecommons.org/licenses/by-sa/4.0/}{{CC-BY-SA-4.0}}. Metadata linking cropped individual specimens to their corresponding group images and NEON identifiers ({\small\texttt{individualID}}) is provided in {\small\texttt{individual\_specimens/metadata.csv}}, while trait measurements are compiled in {\small\texttt{BeetleMeasurements.csv}}. Group images are stored in JPG format, and individual specimens in PNG format. Additional documentation and usage notes are available in {\small\texttt{README.md}} file.

\begin{table}[ht]
\centering
\renewcommand{\arraystretch}{1.05}
\setlength{\tabcolsep}{6pt}
\resizebox{\linewidth}{!}{%
\begin{tabular}{@{}llrl@{}}
\toprule
\textbf{Folder / file} & \textbf{Content} & \textbf{\# records} & \textbf{Licence}\\
\midrule
{\small\texttt{group\_images/}} & Original group images (vial) & \num{577} & CC-BY-SA-4.0\\
{\small\texttt{group\_images\_resized/}} & Resized original group images (vial) & \num{577} & CC-BY-SA-4.0\\
{\small\texttt{individual\_specimens/}} &  Cropped images of individual vial beetles & \num{11654} & CC-BY-SA-4.0\\
{\small\texttt{individual\_specimens/metadata.csv}} & Specimen-level metadata & \num{11654}  & CC-BY-SA-4.0\\
{\small\texttt{BeetleMeasurements.csv}} & Specimen-level trait measurements & \num{39064} & CC-BY-SA-4.0\\
{\small\texttt{Separate\_segmented\_train\_test\_splits\_80\_20/}} & Train/test splits for model training &  — & CC-BY-SA-4.0\\
{\small\texttt{README.md}} & Dataset card and usage notes & — & —\\
\bottomrule
\end{tabular}%
}
\caption{Directory structure of the \emph{vial specimens} repository: \href{https://huggingface.co/datasets/imageomics/2018-NEON-beetles}{\small\texttt{imageomics/2018-NEON-beetles}}.}
\label{tab:2018neonbeetles}
\end{table}

\paragraph{Image Naming.}
Original group images are named using the format {\small\texttt{group\_images/<pictureID>.jpg}}, where {\small\texttt{pictureID}} corresponds to the unique barcode of the vial (e.g., {\small\texttt{group\_images/A00000046094.jpg}}) or, in some cases, the NEON sample ID (e.g., {\small\texttt{MOAB\_001.S.20180724.jpg}}). Resized group images in {\small\texttt{group\_images\_resized/}} follow the same naming convention. Individually cropped specimens are stored under {\small\texttt{individual\_specimens/}} in subfolders ({\small\texttt{part\_000/}}, {\small\texttt{part\_001/}}), each image named as \\
\centerline{{\small\texttt{individual\_specimens/part\_<part\_number>/<pictureID>\_specimen\_<number>.png}}}
Here, {\small\texttt{<pictureID>}} matches the corresponding group image, {\small\texttt{<part\_number>}} (e.g., {\small\texttt{part\_000}} or {\small\texttt{part\_001}}) organizes images into subfolders for manageability, and {\small\texttt{<number>}} is an integer indicating the beetle’s position in the group image, numbered left to right, top to bottom (as if reading a book). 

For the curated machine-learning subset ({\small\texttt{Separate\_segmented\_train\_test\_splits\_80\_20/}}), images are named {\small\texttt{beetle\_<number code>.png}} and organized into {\small\texttt{train/}} or {\small\texttt{test/}} subfolders by species, with metadata linking each image to its species and split.

\paragraph{Metadata.}
Specimen-level metadata are provided in the CSV file {\small\texttt{individual\_specimens/metadata.csv}}, linking individual and group images to the NEON data. The records in this file includes these columns: {\small\texttt{individualImageFilePath}}, {\small\texttt{groupImageFilePath}}, {\small\texttt{NEON\_sampleID}}, {\small\texttt{scientificName}}, {\small\texttt{siteID}}, {\small\texttt{site\_name}}, {\small\texttt{plotID}}, and {\small\texttt{file\_name}}. 

The {\small\texttt{individualImageFilePath}} contains the path to cropped individual specimen images, while {\small\texttt{groupImageFilePath}} links to the parent group image. The {\small\texttt{NEON\_sampleID}} is a unique identifier for the sample, prefixed by the {\small\texttt{plotID}}, enabling linkage to NEON’s broader database. The {\small\texttt{scientificName}} provides the binomial name (\textit{Genus species}), with 78 species across 36 genera, though 10 individuals are labeled only to genus/subgenus, and 17 individuals in {\small\texttt{MOAB\_001.S.20180724.jpg}} lack taxonomic labels (identified as \textit{Cicindela punctulata punctulata} in NEON records). The {\small\texttt{siteID}} and {\small\texttt{site\_name}} indicate the NEON site of collection (30 unique site IDs, 43 site names with some inconsistencies), and {\small\texttt{plotID}} specifies the plot within the site (144 unique plots). The {\small\texttt{file\_name}} supports dataset viewer functionality by providing relative paths to images.

Additional metadata for the {\small\texttt{Separate\_segmented\_train\_test\_splits\_80\_20/}} subset is provided in its corresponding CSV file within the same directory ({\small\texttt{*/metadata.csv}}), with columns: {\small\texttt{filename}} (e.g., {\small\texttt{beetle\_<number code>.png}}), {\small\texttt{md5}} (unique image hash), {\small\texttt{species}} (species epithet), {\small\texttt{split}} (train or test), {\small\texttt{file\_name}} (path for dataset viewer), and {\small\texttt{subset}} (identifier for this subset). Importantly, this subset was developed independently of the present work, as part of a separate dataset processing effort. The images in this subset were generated using image separation and segmentation workflows implemented in the \href{https://huggingface.co/datasets/imageomics/2018-NEON-beetles}{\small\texttt{imageomics/2018-NEON-beetles}} repository. As such, this subset has distinct provenance and citation requirements, distinct from the present study\footnote{If the separate segmented splits subset (images in {\texttt{Separate\_segmented\_train\_test\_splits\_80\_20/}}) is used, please follow the citation guidance in {\texttt{Separate\_segmented\_train\_test\_splits\_80\_20/README}}, which includes citing the dataset and the processing code repository (\href{https://huggingface.co/datasets/imageomics/2018-NEON-beetles}{\texttt{imageomics/2018-NEON-beetles}}) used to produce this subset of images \cite{isadora_e._fluck_2025,Ramirez_2018_NEON_Ethanol-preserved_2025}.}.

\paragraph{Annotations and Trait Measurements.}
The two morphological traits of the vial specimens (elytral length and width) were annotated and accumulated in the CSV file {\small\texttt{BeetleMeasurements.csv}}, with 39,064 measurement records corresponding to elytral length and width measurements for 11,654 individual beetles (measurements of elytral length and width for the same specimen by the same annotator are recorded as two separate entries). Each row in the file includes {\small\texttt{pictureID}} (group image identifier, e.g., {\small\texttt{A00000046094.jpg}}), {\small\texttt{scalebar}} (\textit{x,y} pixel coordinates of a \SI{1.0}{\centi\metre} scalebar), {\small\texttt{cm\_pix}} (pixels per centimeter), {\small\texttt{individual}} (beetle position in the group image, numbered left to right, top to bottom, with a caveat that Notes from Nature resets IDs after 99), {\small\texttt{structure}} ({\small\texttt{ElytraLength}} or {\small\texttt{ElytraWidth}}), {\small\texttt{lying\_flat}} (Yes/No, indicating if the beetle is twisted, affecting width measurements), {\small\texttt{coords\_pix}} (pixel coordinates of elytra measurement on resized images), {\small\texttt{dist\_pix}} (Euclidean distance in pixels), {\small\texttt{dist\_cm}} (distance in centimeters, calculated as {\small\texttt{dist\_pix/cm\_pix}}), {\small\texttt{scientificName}}, {\small\texttt{NEON\_sampleID}}, {\small\texttt{siteID}}, {\small\texttt{site\_name}}, {\small\texttt{plotID}}, {\small\texttt{user\_name}} (annotator’s username), {\small\texttt{workflowID}}, {\small\texttt{genus}}, {\small\texttt{species}}, {\small\texttt{combinedID}} (non-unique due to Notes from Nature limitations), {\small\texttt{measureID}} (unique measurement identifier), {\small\texttt{file\_name}} (path to group image), {\small\texttt{image\_dim}} (full-size image dimensions), {\small\texttt{resized\_image\_dim}} (resized image dimensions), and {\small\texttt{coords\_pix\_scaled\_up}} (coordinates adjusted for full-size images). Measurements were calculated in centimeters using the same calibration factor $k$ used for the pinned specimens. Three annotators performed the measurements (each independently measuring all beetles within an assigned group image in a disjoint manner), with 234 images annotated by all three to evaluate observation error (see details in \flexiref{sec:inter-annotator}{Vial Specimens: Inter-annotator Agreement}).

\mysection{Technical Validation}
\label{validation}
In this section, we present a comprehensive technical validation of the dataset, focusing on experiments and analyses that substantiate its quality and reliability. We evaluated dataset completeness and potential biases through taxonomic and geographic representation, and validated trait measurements by comparing manual and digital methods. Our analyses demonstrate the robustness of the dataset and its suitability for ecological and taxonomic research.

\mysubsection{Dataset Completeness and Bias Analysis}\label{bias-analysis}
\mysubsubsection{Taxonomic and Geographic Representation}\label{species-representation}

\begin{figure}[!t]
\centering
\includegraphics[width=\textwidth]{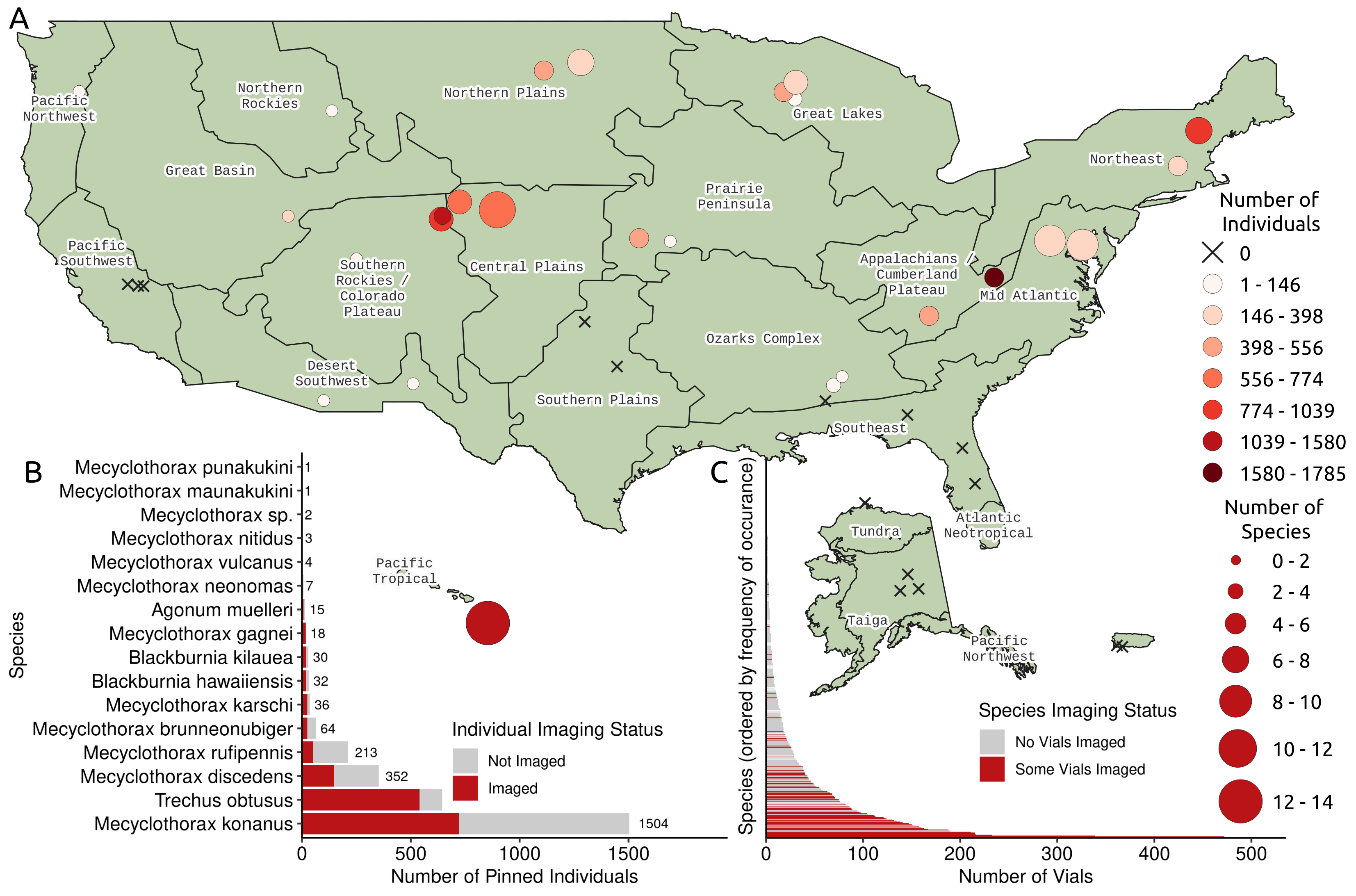}
\captionsetup{justification=justified}
\caption{\textbf{(a)} Map of NEON with points representing each NEON site colored by the number of individual ground beetles at each site and point size scaled by the number of species at each site. \textbf{(b)} Imaging status of all NEON ground beetles to be pinned after sampling at PUUM. \textbf{(c)} Imaging status of species occurring in vials in the 2018 sample record. }
\label{fig:distributionmap}
\end{figure}

Geographically, the dataset includes specimens from 30 of NEON's 47 terrestrial sites, spanning 14 of the 20 ecoclimatic domains (Fig.~\ref{fig:distributionmap}a). The most extensively represented domains include Central Plains (D10; $N\!=\!2913$ specimens), Appalachians and Cumberland Plateau (D07; $N\!=\!2315$), and Northeast (D01; $N\!=\!1398$). Notable geographic gaps exist in the Southeastern United States, the Pacific Southwest, and Alaska. These gaps may be due to fewer abundant taxa in these regions — since at the NEON Biorepository, taxa represented by ten or fewer individuals are pinned, and more abundant ones are preserved in vials.

The dataset contains a substantial portion of NEON's carabid beetle collection, with two complementary components. Its taxonomic composition was validated against existing NEON inventories to quantify the representation of the broader collection. The pinned specimens from the PUUM site include 1579 individuals representing 14 species, constituting a comprehensive collection of the sampled carabid diversity at this Hawaiian site (Fig.~\ref{fig:distributionmap}b). The taxonomic distribution closely matches that of NEON's pinned sampling records for PUUM, with \textit{Mecyclothorax konanus} ($N\!=\!738$), \textit{Trechus obtusus} ($N\!=\!540$), and \textit{Mecyclothorax discedens} ($N\!=\!146$) being the most abundant species. 

The vial specimens comprise 11,654 individuals across 79 different species (15.7\% of species collected in 2019), sourced from 567 of the 6,636 carabid vials (8.5\%) collected by NEON in 2018. While the majority of imaged specimens represent abundant species, the dataset includes representatives across the long-tail distribution of species occurrence (Fig.~\ref{fig:distributionmap}c). The most common genera in this component include \textit{Pterostichus} ($N\!=\!2274$), \textit{Carabus} ($N\!=\!1972$), and \textit{Calathus} ($N\!=\!1599$), collectively representing 51.1\% of the vial specimens.

\mysubsection{Trait Measurements Validation}\label{trait-measure}

\mysubsubsection{Trait Annotation on Pinned Specimens: TORAS on images vs. Calipers on specimens}\label{physical_vs_digital} 
To validate the accuracy of TORAS-based digital measurements and establish ground truth values, we conducted parallel measurements on a subset of specimens physically using high-precision calipers. A randomly selected subset of 64 beetle specimens from the pinned beetles was used for this validation. We aimed to select 6 specimens per species, capturing a range of specimen positioning conditions relevant to training measurements: 2 well-positioned (i.e., aligned straight, parallel to the background surface, and with closed elytra), 2 poorly positioned (e.g., visibly tilted and/or with open elytra), and 2 intermediate (e.g., slightly tilted and/or with slightly opened elytra). Due to differences in species abundance, some rarer species had fewer than 6 available specimens, resulting in a total of 64 individuals for 14 species.

\begin{figure}[!t]
\centering
\includegraphics[width=\textwidth]{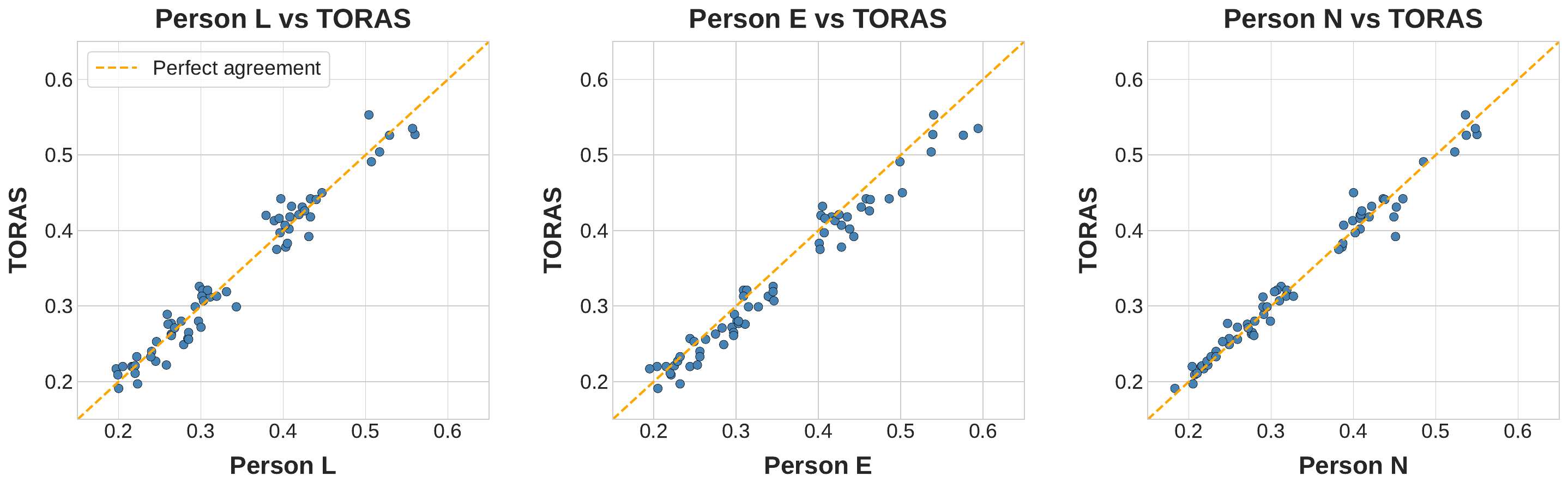}
\captionsetup{justification=justified}
\caption{Scatter plot comparing the elytral length measurements (in \SI{}{\centi\metre}) on \textit{pinned specimens} between three human annotators (labeled as Person L, Person E and Person N) using calipers on physical specimens and the TORAS system on beetle images. The x-axis represents TORAS measurements, while the y-axis shows the calipers measurements. The orange dashed line denotes the line of perfect agreement ($y = x$). Data points are color-coded in blue, revealing a tight clustering around the line of agreement and thus demonstrating excellent concordance between the TORAS and calipers measurements.}
\label{fig:TORAS-Calipers}
\end{figure}

\begin{figure}[!t]
\centering
\includegraphics[width=\textwidth]{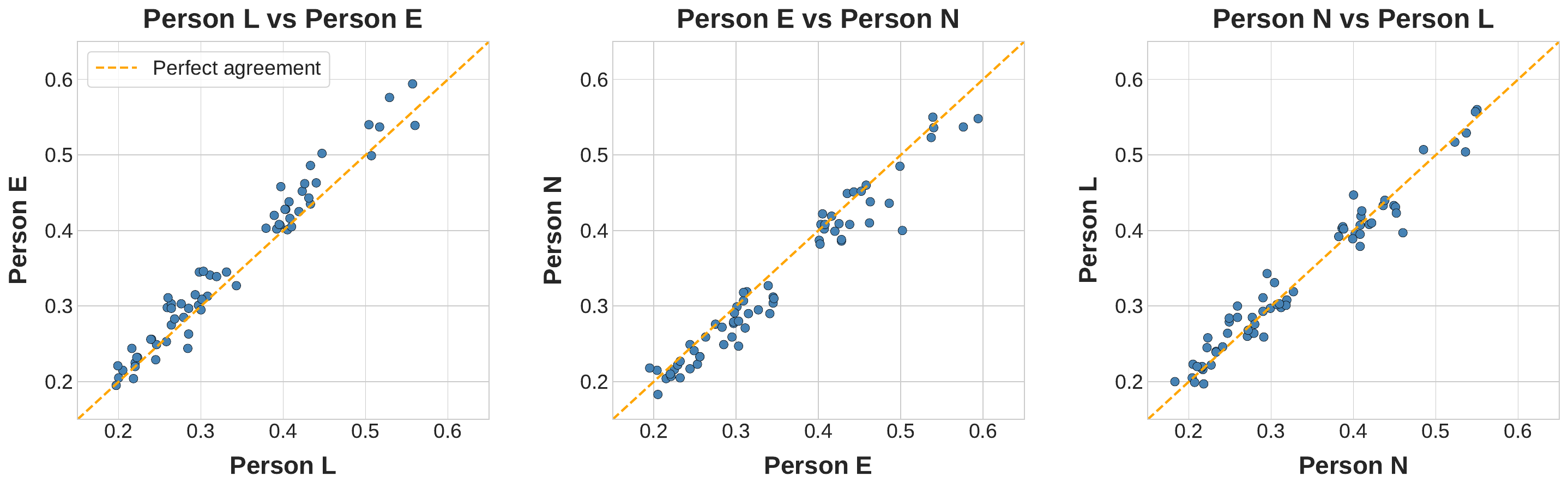}
\captionsetup{justification=justified}
\caption{Scatter plot depicting inter-annotator agreement for elytral length measurements (in \SI{}{\centi\metre}) on \textit{pinned specimens} among three human annotators (Person L, Person E, and Person N) using digital calipers on physical specimens. The figure consists of three subplots, each comparing a pair of annotators: Person L vs. Person E, Person E vs. Person N, and Person N vs. Person L. In each subplot, the orange dashed line indicates perfect agreement, and data points are shown in blue, illustrating the level of consistency and variability in manual measurements across observers.}
\label{fig:IAA-Pinned}
\end{figure}

Three annotators measured the morphological traits (elytral length, basal pronotum width, and maximum elytral width) on each specimen. The multi-observer approach was implemented to minimize human error and assess measurement consistency. Each annotator conducted their measurements independently, without access to the other's results or the corresponding TORAS-derived measurements. All measurements were performed using digital calipers with 0.01 mm precision, under standardized lighting conditions. Upon completion, the measurements were compared against TORAS-derived measurements (See \cref{fig:TORAS-Calipers}, \cref{tab:pinned-all-stats}a) and among annotators (See \cref{fig:IAA-Pinned}, \cref{tab:pinned-all-stats}b) to evaluate consistency and accuracy.

\begin{table}[!t]
\centering
% \begin{tabular}{cccc cccc}
\begin{tabular}{cccc@{\hspace{2.5em}}cccc}
\toprule
\multicolumn{4}{c}{\textbf{(a) Calipers vs TORAS}} & \multicolumn{4}{c}{\textbf{(b) Inter-Annotator}} \vspace{.2em} \\
\textbf{Annotation Pair} & \textbf{RMSE} & \textbf{R\textsuperscript{2}} & \textbf{Bias} &
\textbf{Annotation Pair} & \textbf{RMSE} & \textbf{R\textsuperscript{2}} & \textbf{Bias} \\
\midrule
Person L v TORAS & $0.0199$ & $0.9586$ & $\phantom{+}0.0011$ & Person L v Person E & $0.0255$ & $0.9389$ & $-0.0156$ \\
Person E v TORAS & $0.0255$ & $0.9321$ & $-0.0167$ & Person E v Person N & $0.0277$ & $0.9246$ & $\phantom{+}0.0175$ \\
Person N v TORAS & $0.0153$ & $0.9755$ & $-0.0008$ & Person N v Person L & $0.0201$ & $0.9570$ & $-0.0019$ \\
\midrule
Mean (L, N, E) v TORAS & $0.0150$ & $0.9765$ & $\phantom{+}0.0057$ &
IA-Average & $0.0244$ & $0.9402$ & $\phantom{+}0.0116$ \\
\bottomrule
\end{tabular}
\captionsetup{justification=justified}
\caption{Statistical metrics assessing \textbf{(a)} the agreement between human annotators using digital calipers and TORAS on measuring elytral length on \textit{pinned specimens}; and (b) the agreement among the three human annotators. The table presents root mean square error (RMSE in cm), coefficient of determination (R\textsuperscript{2}), and bias (in cm) for each pairwise comparison and averages. The \textit{`Mean (L,N,E) v TORAS'} row represents metrics computed between the averaged calipers measurements (mean of measurements from Persons L, N and E) and TORAS, while the \textit{`IA-Average'} row refers to the mean values of each metric, computed as the average of the absolute values of the pairwise inter-annotator results.\protect\footnotemark}
\label{tab:pinned-all-stats}
\end{table}

\footnotetext{%
\(\displaystyle
\text{IA-Average} = \frac{1}{3} \sum_{i=1}^{3} \big| M_i \big|,
\)
where \(M_i\) represents the metric (RMSE, R\textsuperscript{2}, or Bias) for each of the three inter-annotator pairs: (L,E), (E,N), and (N,L).
}

\paragraph{Analysis.} The scatter plots in Figure \ref{fig:TORAS-Calipers} compare the elytra length measurements between TORAS and each of the three human annotators using calipers. Each subplot corresponds to the pair between one annotator and TORAS, while the diagonal orange dashed line represents perfect agreement where calipers and TORAS measurements would be identical ($y=x$). Across all three subplots, data points cluster closely along this line, indicating that TORAS captures elytral length with high fidelity across annotators. Minor deviations reflect individual measurement tendencies rather than systematic bias, as corroborated by the near-zero bias values and high R\textsuperscript{2} scores reported in Table \ref{tab:pinned-all-stats}. This close correspondence validates the accuracy of TORAS-derived measurements against the ground truth established by calipers measurements, confirming that TORAS can reliably measure the morphological traits under varying specimen positioning conditions. To further quantify the agreement between human annotators and TORAS, we calculated three statistical metrics: root mean square error (RMSE), coefficient of determination (R\textsuperscript{2}), and bias. RMSE measures the average magnitude of differences between measurements, with lower values indicating better precision. R\textsuperscript{2} assesses the strength of the linear relationship, where values closer to 1 signify that one set of measurements explains most of the variance in the other. Bias quantifies systematic differences, with values near zero suggesting no consistent over- or underestimation. 

The visual pattern in \cref{fig:TORAS-Calipers} is quantitatively supported by the statistics in Table \ref{tab:pinned-all-stats}, where RMSE values for individual annotators range from $0.0153$ to $0.0255$ cm, R\textsuperscript{2} values from $0.9321$ to $0.9755$, and biases remain close to zero ($-0.0167$ to $0.0011$ cm). Among these, Person N exhibits the closest alignment with TORAS (RMSE $0.0153$ cm and R\textsuperscript{2} $0.9755$), reflecting highly consistent measurements. When TORAS is compared to the average of all three measurements (i.e., mean of Person L, N, and E), the performance metrics improve further (RMSE $0.0150$ cm and R\textsuperscript{2} $0.9765$, and bias $0.0057$ cm). 

Furthermore, we evaluated the consistency among human annotators' measurements to understand the natural variability in manual measurements. Figure \ref{fig:IAA-Pinned} presents the inter-annotator agreements for elytral length measurements, with each subplot comparing a pair of annotators. The data points cluster around the line of perfect agreement, though with slightly greater dispersion than observed in the TORAS–calipers comparisons (\cref{fig:TORAS-Calipers}), reflecting the inherent variability and subjectivity of manual measurement even under standardized conditions.

Quantitatively, inter-annotator comparisons show a higher average RMSE ($0.0244$ cm) and bias ($0.0000$ cm), and lower average R\textsuperscript{2} ($0.9402$), scores relative to the TORAS–human results (Table \ref{tab:pinned-all-stats}). These patterns indicate that TORAS attains accuracy comparable to, and in certain cases exceeding, that of human annotators. The lower RMSE and higher R\textsuperscript{2} values for TORAS, relative to inter-human variability, suggest that the system not only replicates human precision but also mitigates subjective inconsistencies inherent to manual measurements. The consistent near-zero bias across all comparisons (See \cref{tab:pinned-all-stats}a, Column 3) further supports the absence of systematic deviation. Collectively, these findings demonstrate that TORAS delivers measurements that are both reliable and reproducible, validating its effectiveness as an automated tool for large-scale morphological trait extraction from digital images of beetle specimens.

\mysubsubsection{Trait Annotation for Vial Specimens: Inter-Annotator Agreement}\label{sec:inter-annotator}
Similar to the analysis conducted on \textit{pinned specimens}, we evaluated the reliability and consistency of trait measurements on \textit{vial specimens} by comparing independent measurements from three annotators. Each annotator measured a subset of images from the vial specimens, with 234 images annotated by all three to assess observation error and inter-annotator variability. The analysis focused primarily on elytral length, enabling a direct assessment of measurement consistency and potential biases under the same standardized annotation protocol used for pinned specimens.

\begin{figure}[!t]
\centering
\includegraphics[width=\textwidth]{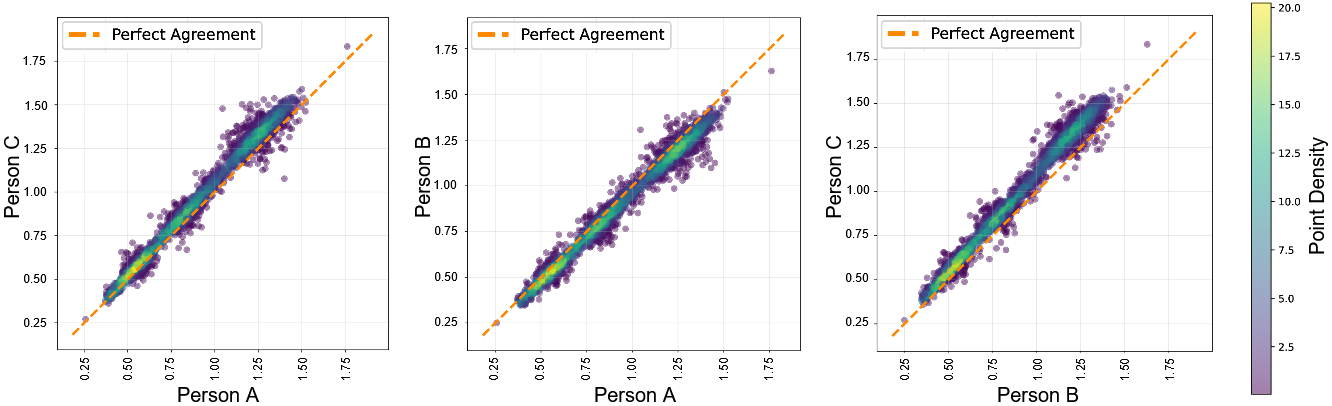}
\captionsetup{justification=justified}
\caption{Pairwise scatter-density plots illustrating inter-annotator agreement for elytral length measurements of \textit{vial specimens} across three independent annotators (Person A, Person B, and Person C). Each plot compares the measurements (in centimeter) of one annotator against another, with the color gradient representing point density (darker colors indicate higher density). The dashed orange line represents perfect agreement, showing a tight clustering of data points along the diagonal, indicating high consistency among annotators.}
\label{fig:inter_annotator_aggrement}
\end{figure}

\paragraph{Analysis.} The pairwise scatter-density plots in Figure~\ref{fig:inter_annotator_aggrement} illustrate the agreement in elytral length measurements among the three annotators. Each subplot represents a distinct annotator pair, with the orange dashed diagonal ($y=x$) indicating perfect correspondence. In all three subplots, the data points cluster tightly along the diagonal $y=x$ line. This near-perfect alignment with minimal dispersion (i.e. random scatter around the diagonal line in each subplot) indicates a high degree of precision and strong concordance, and reveals negligible systematic bias among the annotators across the entire size range of the vial specimens, mirroring the consistency observed for pinned specimens.

Quantitative results, summarized in \cref{tab:inter_annotator_stats}b, were computed using the same three statistical metrics: RMSE, R\textsuperscript{2}, and bias. The average RMSE of $0.0752$ reflects limited variability across annotators, while the mean R\textsuperscript{2} of $0.9356$ demonstrates a strong linear relationship among measurements. The small average bias of $-0.0336$ suggests no consistent over- or underestimation trend. These findings demonstrate a high degree of inter-annotator consistency and validate the reliability of measurements for subsequent large-scale trait analyses.

\begin{table}[!t]
\centering
\begin{tabular}{cccc}
\toprule
\textbf{Annotator Pair} & \textbf{RMSE} & \textbf{R\textsuperscript{2}} & \textbf{Bias} \\
\midrule
Person A vs Person B & 0.0553 & 0.9687 & $\phantom{+}0.0374$ \\
Person A vs Person C & 0.0684 & 0.9521 & $-0.0503$ \\
Person B vs Person C & 0.1020 & 0.8859 & $-0.0878$ \\
\midrule
Average
& 0.0752 & 0.9356 & $-0.0336$ \\
\bottomrule
\end{tabular}
\captionsetup{justification=justified}
\caption{Inter-annotator agreement for elytral length measurements of \textit{vial specimens} among Person A, Person B, and Person C. We have reported the root mean square error (RMSE), coefficient of determination (R\textsuperscript{2}), and bias for each pairwise comparison, along with their averages across all pairs.}
\label{tab:inter_annotator_stats}
\end{table}

\mysubsubsection{Trait Measurement Tools: TORAS vs Notes from Nature} 
The selection between TORAS and Notes from Nature for trait measurement depends on the specific characteristics of our specimen types and the capabilities of each platform. For pinned specimens, we used TORAS, which offers high precision through its AI-powered annotation tools and supports high-resolution images without compression, ensuring accurate morphological measurements. For vial specimens, we used Notes from Nature, which allows multiple participants to measure the same specimens, significantly reducing the time required compared to traditional methods. However, it requires image resizing due to upload size limitations, which reduces pixel resolution and accounts for minor variations in measurements among participants. Additionally, its indexing system poses a challenge, as annotation indices for a single image reset from 100 to 1, resulting in duplicate index numbers for specimens beyond the 100th. To address this, we implemented a coordinate-based disambiguation system during post-processing for group images containing more than 100 beetles, ensuring unique specimen identifiers. The complementary use of these platforms leverages the precision of expert-driven measurements for detailed morphological analysis while taking advantage of Notes from Nature’s efficiency for large-scale trait data collection.

\begin{figure}[!t]
\centering
\includegraphics[width=\textwidth]{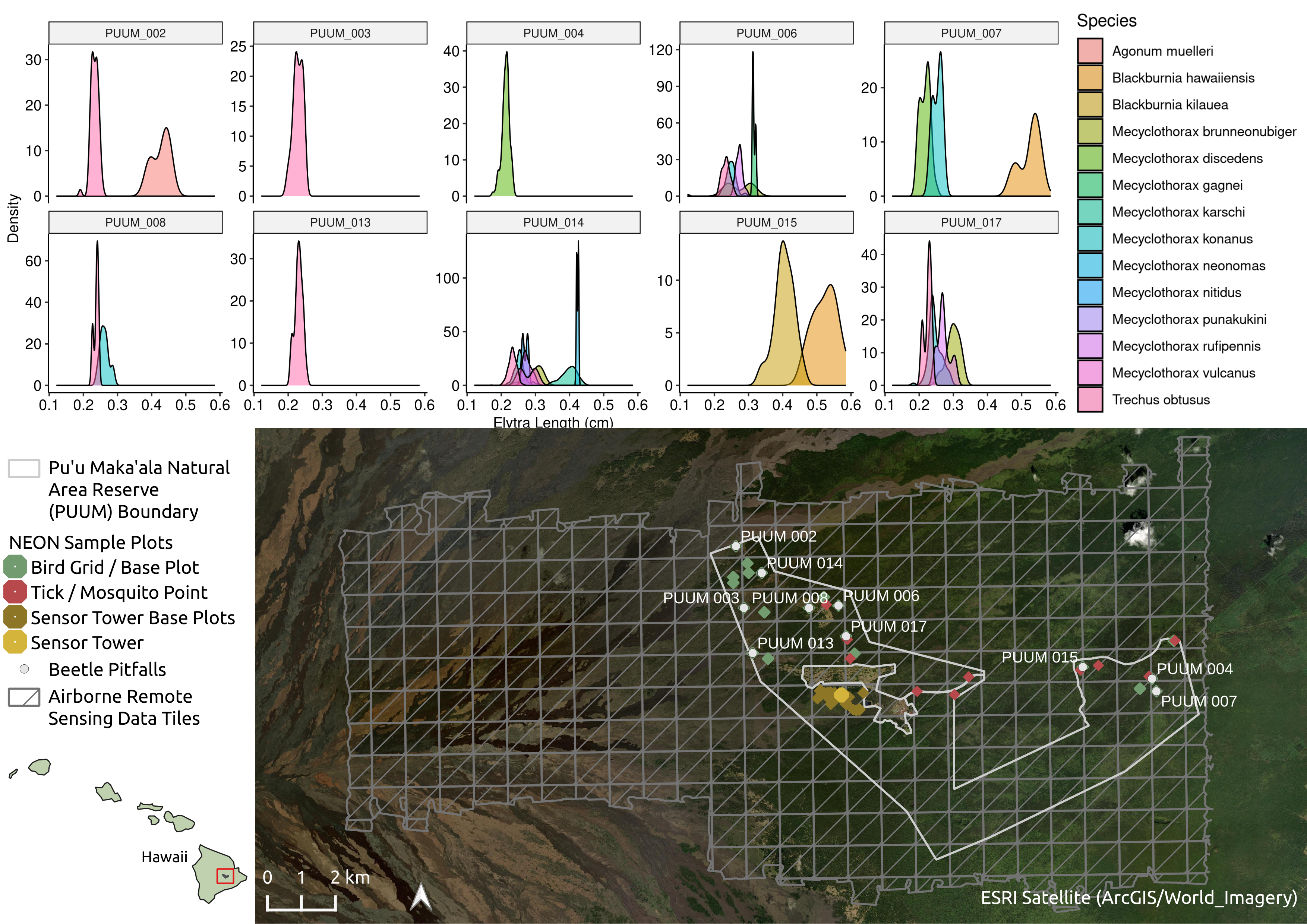}
\captionsetup{justification=justified}
\caption{Density distributions of elytral lengths of pinned specimens across all sample plots at the PUUM site. Beetle body size measurements can be collocated with other NEON Data streams, including Bird Grids and Vegetation Base Plots (green), Airborne Remote Sensing Data (grey tiles), and in proximity to automated sensor data associated with the Flux Tower (yellow).}
\label{fig:Bodysize}
\end{figure}

\mysection{Usage Notes}\label{usage}
This section provides technical guidance for researchers reusing the data, with a focus on integrating it with NEON's broader data infrastructure and leveraging it for interdisciplinary applications. Pinned specimens link to NEON by {\small\texttt{IndividualID}} to the {\small\texttt{bet\_parataxonomistID}} and {\small\texttt{bet\_expertTaxonomistIDProcessed}} data tables, vials link with {\small\texttt{sampleCode}} to the {\small\texttt{bet\_sorting}} and associated downstream tables \cite{DP1.10022.001_RELEASE2025}. The linkage between these images and the broader NEON database enables a more refined view of the multimodal data in this dataset. With the association between {\small\texttt{IndividualID}} or {\small\texttt{sampleCode}}, images and annotations can be spatially rectified, and queried against NEON's other data streams (Figure~\ref{fig:Bodysize}).

For researchers working with computer vision or machine learning, this dataset offers unique opportunities to develop multimodal models. For example, NEON's \textit{Airborne Observation Platform (AOP)} data, which includes high-resolution imaging spectroscopy, LiDAR, and RGB imagery, can be combined with ground-based beetle observations to study relationships between spectral signatures, habitat structure, and species distributions. The standardized sampling design and long-term nature of NEON data further support temporal analyses and the training of predictive models for ecological change.

To facilitate data integration, users can leverage the {\small\texttt{neonUtilities}} R package to merge site and month-specific files into unified datasets. The {\small\texttt{geoNEON}} package provides tools for precise geolocation of sampling plots, enhancing spatial analyses. For taxonomic updates or corrections, the {\small\texttt{bet\_identificationHistory}} table tracks revisions, ensuring transparency in species identifications over time \cite{DP1.10022.001_RELEASE2025}.

For ecologists, these data enable investigations into biodiversity patterns across 14 of NEON's 20 ecoclimatic Domains, species responses to climate gradients, and trophic interactions through co-located data on plants, soils, and microclimate. The long-term, spatially replicated sampling design (6 plots per site since 2023) -- available for the pinned specimens from Hawaii in this dataset -- supports studies of phenological shifts, population dynamics, and habitat associations, particularly when combined with NEON's land cover classifications and sensor-derived environmental data.  Studies using similarly structured datasets from NEON have already demonstrated the potential of individual-level functional trait data to reveal biodiversity/ecological patterns across multiple environmental gradients. For example, \iftoggle{arxiv}{\citet{read2018among}}{Read et al. (2018)\cite{read2018among}} used individual body size measurements of small mammals collected across the NEON sites to test niche-based theories of community assembly, revealing that warmer environments support greater interspecific trait differentiation and expanded niche space. Likewise, \iftoggle{arxiv}{\citet{mcgrew_abiotic_nodate}}{McGrew et al. (2021)\cite{mcgrew_abiotic_nodate}} analyzed fish body size data from the NEON freshwater sites and showed how abiotic variables are key for structuring both intraspecific trait variation and species richness. Such studies highlight the potential of individual-level trait datasets to uncover the mechanisms linking environment, trait distributions and biodiversity patterns across scales. Our study extends this potential to invertebrates by providing a unique, large scale dataset of carabids' body sizes.

\mysubsection{Code Availability}
All code developed and used in this study is publicly available to support transparency and reproducibility. This includes implementations for (i) detecting and localizing individual specimens in group images using Grounding-DINO; (ii) analyzing species-level and plot-level distribution patterns; (iii) computing density distributions of elytral lengths for pinned specimens across all sampling plots at the PUUM site; (iv) assessing inter-annotator agreement; and (v) validating TORAS-based morphometric measurements against manual caliper measurements. The complete, well-documented codebase, along with usage instructions and dependencies, is hosted on GitHub and described in detail in the repository ({\small\texttt{README.md}} file): \href{https://github.com/Imageomics/carabidae_beetle_processing}{\small\texttt{github.com/Imageomics/carabidae\_beetle\_processing}} \cite{Rayeed_Carabidae_Beetle_Processing_2025}.

\mysubsection{Acknowledgment}
This work was supported by both the Imageomics Institute and the AI and Biodiversity Change (ABC) Global Center. The Imageomics Institute is funded by the US National Science Foundation's Harnessing the Data Revolution (HDR) program under Award No. 2118240 (Imageomics: A New Frontier of Biological Information Powered by Knowledge-Guided Machine Learning). The ABC Global Center is funded by the US National Science Foundation under Award No. 2330423 and Natural Sciences and Engineering Research Council of Canada under Award No. 585136. This work draws on research funded by the Social Sciences and Humanities Research Council. S. R. and A. E. were additionally supported by the US National Science Foundation's Award No. 242918 (EPSCOR Research Fellows: NSF: Advancing National Ecological Observatory Network-Enabled Science and Workforce Development at the University of Maine with Artificial Intelligence) and by Hatch project Award \#MEO-022425 from the US Department of Agriculture’s National Institute of Food and Agriculture. R.P.G. gratefully acknowledges support from the National Science Foundation (NSF), and in particular NSF DBI \#2027234 that has supported work on Notes on Nature. This work draws on research supported by the National Ecological Observatory Network (NEON), a program sponsored by the National Science Foundation and operated under cooperative agreement by Battelle. This material uses specimens and/or samples collected as part of the NEON Program and is based in part upon work supported by NEON. Additionally, this material is based upon work supported by the National Science Foundation under Award Numbers 2301322, 1950364, and 1926569. This publication uses data generated via the \href{https://www.zooniverse.org/}{\small\texttt{Zooniverse}} platform, the development of which is funded by generous support, including a Global Impact Award from Google, and by a grant from the Alfred P. Sloan Foundation.

\mysubsection{Author Contributions}
S.M.R., S.R., A.E., and M.K. conceived the study and contributed to its conceptualization, data curation, and formal analysis. S.M.R., A.E., M.K., and I.E.F. led dataset review and preparation of the initial manuscript draft. S.R., G.W.T., and B.B. contributed to the investigation and the provision of resources. E.G.C. and M.W.D. performed validation and assisted with data visualization. S.M.R., A.E., M.K., I.E.F., G.W.T., E.G.C., and S.S. developed and implemented the software and analytical workflows. C.V.S., S.C.L., T.B.-W., and P.M. provided supervision, project oversight, and guidance throughout the study. E.D., M.R., A.K., H.L., and J.W. contributed to data acquisition and review. R.P.G. and all other authors contributed to manuscript revision and critical editing. All authors read and approved the final version of the manuscript.

\mysubsection{Competing Interests}
The authors declare no competing interests.




% \end{appendices}

% \clearpage

\bibliography{sn-bibliography}
% if required, 
% the content of .bbl file can be included here
% once bbl is generated
% \input sn-article.bbl

\end{document}